\documentclass[11pt]{article}

\usepackage[final]{latex/acl}

\usepackage{times}
\usepackage{latexsym}

\usepackage[T1]{fontenc}

\usepackage[utf8]{inputenc}

\usepackage{microtype}

\usepackage{inconsolata}

\usepackage{graphicx}

\usepackage{amsthm}              
\usepackage{amssymb}             
\usepackage{amsmath}
\newtheorem{definition}{Definition} 

\usepackage{xcolor}

\usepackage{subcaption}
\usepackage{graphicx}
\usepackage{algorithm}
\usepackage{algorithmic}
\usepackage{amsmath} 

\usepackage{booktabs}     
\usepackage{multirow}     
\usepackage{tabularx}     
\usepackage[table]{xcolor} 
\usepackage{arydshln}     
\usepackage{makecell}     

\usepackage{url}

\title{IDPruner: Harmonizing Importance and Diversity in Visual Token Pruning for MLLMs}

\author{
 \textbf{Yifan Tan\textsuperscript{1,2}},
 \textbf{Yifu Sun\textsuperscript{2}},
 \textbf{Shirui Huang\textsuperscript{2}},
 \textbf{Hong Liu\textsuperscript{2}},
\\
 \textbf{Guanghua Yu\textsuperscript{2}},
 \textbf{Jianchen Zhu\textsuperscript{2}},
 \textbf{Yangdong Deng\textsuperscript{1}}
\\
\\
 \textsuperscript{1}School of Software, Tsinghua University,
 \textsuperscript{2}Tencent
}

\begin{document}
\maketitle

\begin{abstract}

Multimodal Large Language Models (MLLMs) have demonstrated impressive capabilities, yet they encounter significant computational bottlenecks due to the massive volume of visual tokens.
Consequently, visual token pruning, which substantially reduces the token count, has emerged as a critical technique for accelerating MLLM inference.
Existing approaches focus on token importance, diversity, or an intuitive combination of both, without a principled framework for their optimal integration.
To address this issue, we first conduct a systematic analysis to characterize the trade-off between token importance and semantic diversity.
Guided by this analysis, we propose the \textbf{I}mportance and \textbf{D}iversity Pruner (\textbf{IDPruner}), which leverages the Maximal Marginal Relevance (MMR) algorithm to achieve a Pareto-optimal balance between these two objectives.
Crucially, our method operates without requiring attention maps, ensuring full compatibility with FlashAttention and efficient deployment via one-shot pruning.
We conduct extensive experiments across various model architectures and multimodal benchmarks, demonstrating that IDPruner achieves state-of-the-art performance and superior generalization across diverse architectures and tasks.
Notably, on Qwen2.5-VL-7B-Instruct, IDPruner retains 95.18\% of baseline performance when pruning 75\% of the tokens, and still maintains 86.40\% even under an extreme 90\% pruning ratio.
Our code is available at \url{https://github.com/Tencent/AngelSlim}.

\end{abstract}

\begin{figure}[t] 
    \centering
    \includegraphics[width=\linewidth]{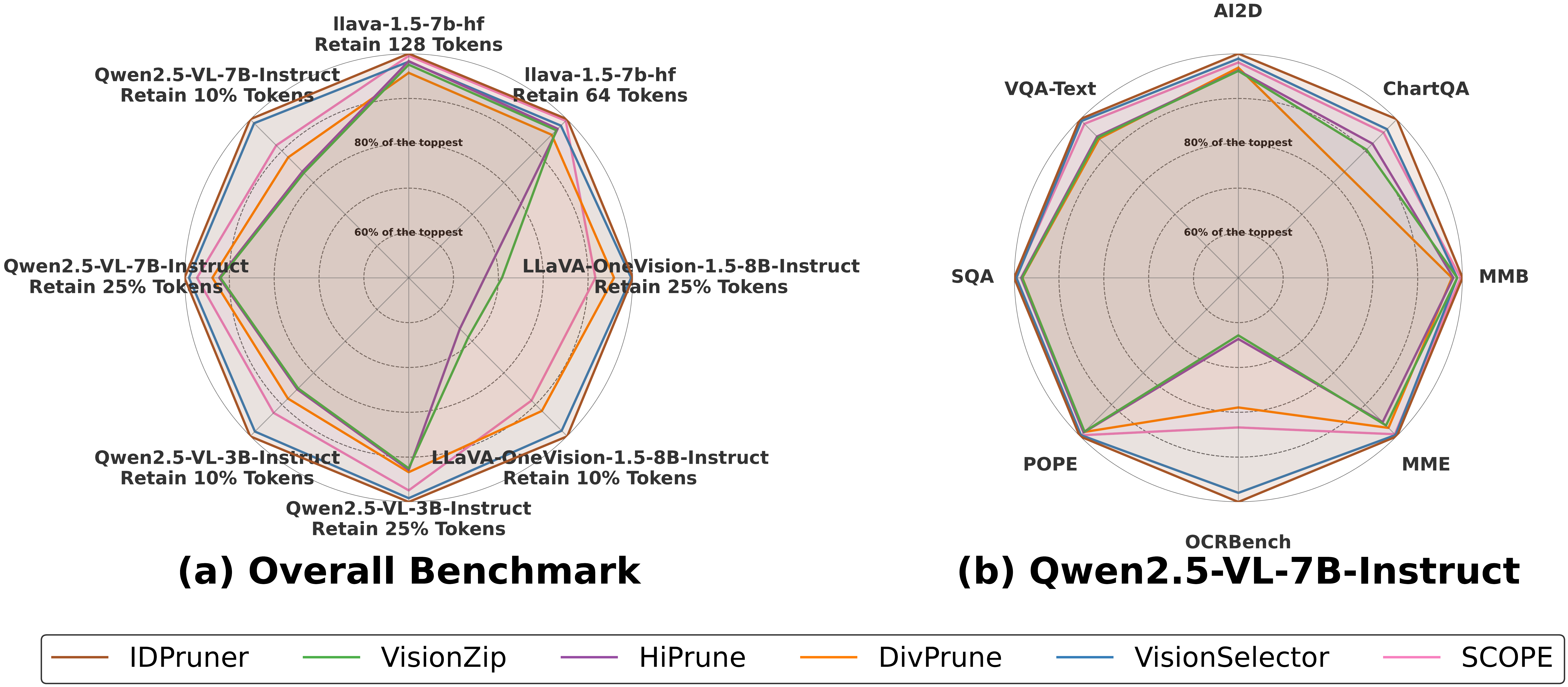} 
    \caption{
        \textbf{Performance comparison across four architectures and eight benchmarks.}
        IDPruner (outermost boundary) consistently outperforms baselines in both \textbf{(a) aggregated performance} across four diverse MLLM architectures and \textbf{(b) fine-grained benchmark breakdown} for Qwen2.5-VL. This demonstrates the superior cross-architecture generalization and task-specific robustness of our method.
    }
    \label{fig:radar_comparison}
\end{figure}

\begin{figure*}[t]
    \centering
    \includegraphics[width=0.95\textwidth]{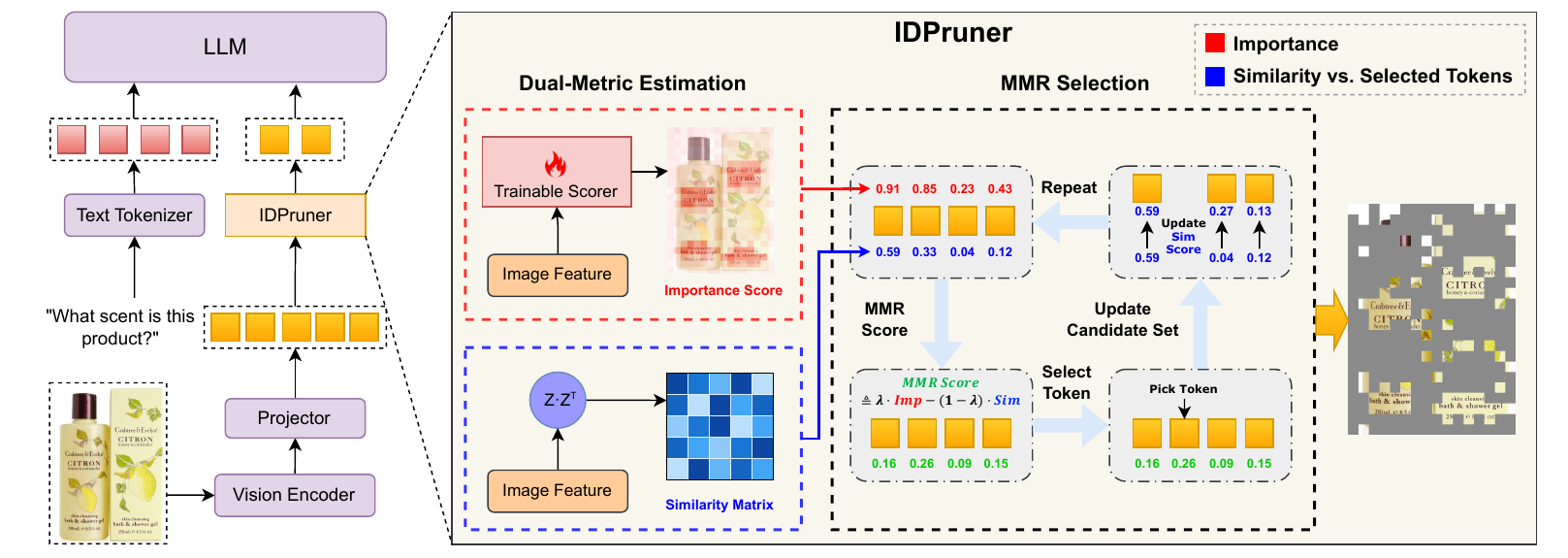} 
    \caption{
        \textbf{Overview of the IDPruner framework.}
        \textbf{Left}: Integration of our one-shot visual token pruning into the MLLM inference pipeline.
        \textbf{Right}: The core mechanism computes \textbf{Importance Scores} (Red) and a \textbf{Similarity Matrix} (Blue), utilizing an MMR selection process to harmonize importance and diversity.
        This approach operates without attention maps and remains compatible with FlashAttention.
    }
    \label{fig:framework}
\end{figure*}

\section{Introduction}
Multimodal Large Language Models (MLLMs) have achieved significant success in artificial intelligence.
These models typically encode images or videos into sequences of visual tokens, which are then processed together with textual inputs by the language model to generate text responses \cite{liu2023llava, liu2023improvedllava}.
For instance, Qwen2.5-VL generates approximately 2,691 visual tokens when processing a single 1080p image (1920×1080), with each token representing a 28×28 pixel patch.
The high number of visual tokens creates a heavy computational burden, limiting the efficiency and practical deployment of MLLMs \citep{Zhou2024ASO}.
Thus, visual token pruning \cite{wang2025effivlmbenchcomprehensivebenchmarkevaluating,Shao2025WhenTT}, which aims to reduce the number of visual tokens while maintaining model performance, has emerged as a critical technique for achieving efficient MLLM inference.

Existing pruning strategies generally fall into two categories: importance-based and diversity-based methods.
Importance-based approaches \cite{chen2024image,yang2025visionzip,yang2025topv} select salient tokens, focusing on foreground objects, but often sacrificing the background context essential for global reasoning.
In contrast, diversity-based methods \cite{Alvar2025DivPruneDV, Zou2025DontJC} maximize semantic coverage to reduce redundancy but risk retaining task-irrelevant noise while missing fine-grained details.
Recent hybrid approaches \cite{Zhang2024BeyondTA, Zhang2025BeyondAO, Li2025Why1} attempt to combine these complementary criteria but lack rigorous analysis, relying on intuition-based integration that yields suboptimal performance.
Therefore, a systematic analytical framework is needed to characterize the interaction between importance and diversity and derive optimal integration strategies.

To address this, we first conduct a systematic analysis to investigate the trade-off between token importance and semantic diversity.
As shown in Figure \ref{fig:simulation_tradeoff}, our analysis reveals that current approaches fail to effectively balance these two critical dimensions.
To overcome this limitation, we introduce the \textbf{I}mportance and \textbf{D}iversity Pruner (\textbf{IDPruner}), a novel pruning strategy designed to balance these criteria optimally.
Specifically, as illustrated in Figure \ref{fig:framework}, we cast visual token pruning as a re-ranking problem in information retrieval and adapt the Maximal Marginal Relevance (MMR) \cite{Carbonell1998TheUO} algorithm to model the interplay between token importance and semantic diversity explicitly.
This approach selects tokens that jointly maximize both importance and diversity.

IDPruner achieves state-of-the-art performance, as demonstrated by comprehensive evaluations on multimodal benchmarks.
Notably, on the Qwen2.5-VL-7B-Instruct model, even under an extreme compression ratio of 90\%, our method retains \textbf{86.40\%} of the baseline performance, significantly outperforming existing competitive approaches.
Crucially, unlike progressive pruning strategies that dynamically change sequence lengths, IDPruner performs one-shot pruning at an early stage, which makes it easier to integrate into inference engines like vLLM \cite{kwon2023efficient}.
Furthermore, our method works without requiring attention information, ensuring full compatibility with FlashAttention \cite{dao2022flashattention} to maximize inference efficiency.

The main contributions of this work are summarized as follows:
\begin{itemize}
    \item We conduct a systematic analysis to characterize the trade-off between token importance and semantic diversity, providing a theoretical basis for their integration.
    \item We propose IDPruner, which adapts the Maximal Marginal Relevance (MMR) algorithm to visual token pruning, enabling the optimal harmonization of importance and diversity.
    \item Extensive experiments demonstrate that our method achieves state-of-the-art performance and exceptional cross-architecture generalization, as visualized in Figure \ref{fig:radar_comparison}, while supporting one-shot pruning and FlashAttention acceleration, offering a practical solution for efficient MLLM deployment.
\end{itemize}

\section{Related work}

\noindent \textbf{Large Multimodal Models and Visual Token Pruning.} Recent Multimodal Large Language Models (MLLMs) \citep{liu2023improvedllava,wang2024qwen2,zhu2025internvl3} have demonstrated impressive capabilities across various visual tasks, yet they encounter significant computational bottlenecks due to the massive volume of visual tokens.
Static-resolution models like LLaVA-1.5 \citep{liu2023improvedllava} and LLaVA-NeXT \citep{liu2024llavanext} require 576 and 2,880 input tokens per image, respectively, while newer architectures such as the Qwen-VL \citep{bai2025qwen2}, LLaVA-OneVision \cite{Li2024LLaVAOneVisionEV}, and InternVL \cite{zhu2025internvl3} series demand comparable token budgets for high-resolution processing.
Consequently, visual token pruning, which eliminates unnecessary tokens, has emerged as a crucial technique for accelerating MLLM inference.
Current research typically falls into two categories: importance-based methods and diversity-based methods.

\noindent \textbf{Importance-based Token Pruning.} Importance-based approaches reduce computational overhead by retaining only the most salient tokens.
Early studies rely on attention scores from LLM decoder layers \citep{chen2024image,zhang2024sparsevlm,xing2024pyramiddrop,zhang2025adaptinfer,ye2025atp,han2025adav}, while subsequent research discovers that the attention of the [CLS] token in Vision Transformers (ViT) provides a more effective importance measure \citep{yang2025visionzip,liu2025hiprune,zhang2024cls,Tong2025FlowCutRR}.
To mitigate limitations such as FlashAttention incompatibility, recent work has introduced alternative metrics, including optimal transport and L2 norms \citep{yang2025topv,Dhouib2025PACTPA}.
Beyond training-free methods, approaches like VisionSelector \cite{zhu2025visionselectorendtoendlearnablevisual} employ learnable modules to estimate token importance, achieving state-of-the-art performance through end-to-end training.
Despite their effectiveness in capturing region-specific details, these methods often overlook global context, potentially causing information loss in background areas.

\noindent \textbf{Diversity-based Token Pruning.} Diversity-based approaches aim to preserve information coverage by regarding visual tokens as a collective set, minimizing redundancy to retain a representative subset of visual features.
DivPrune \cite{Alvar2025DivPruneDV} formulates this task as a Max-Min Diversity Problem, solving it via a greedy algorithm to maximize semantic coverage, while DART \cite{Wen2025StopLF} employs a parallelizable strategy that selects pivot tokens and eliminates their nearest neighbors to maintain diversity.
However, maximizing redundancy reduction often comes at the cost of missing fine-grained details in focal regions, as these methods may indiscriminately retain task-irrelevant noise.

\noindent \textbf{Hybrid Strategies.} Synergizing importance and diversity typically yields superior performance compared to single-criterion methods.
VisPrune \cite{Zhang2024BeyondTA} allocates token budgets based on both [CLS] attention and diversity, while CDPruner \cite{Zhang2025BeyondAO} employs Determinantal Point Processes (DPP) to balance these objectives.
Other approaches explore alternative integration strategies, such as ensuring spatial coverage via region-based allocation \cite{Zou2025DontJC, Arif2025HiREDAT} or modeling pruning as a set cover problem to optimize multimodal coverage \cite{Li2025Why1, Deng2025SCOPESO}.
Although effective, these methods typically rely on heuristic integration strategies without a systematic analytical framework.
In this work, we address this limitation by introducing a systematic framework that optimally harmonizes importance and diversity.

\section{An Empirical Analysis of the Importance-Diversity Trade-off}
\label{sec:analysis}

\subsection{Quantifying Importance and Diversity}

Visual token pruning strategies typically focus on either importance-based selection or diversity preservation; however, balancing these two goals remains challenging.
To systematically analyze the relationship between these two paths, we first reformulate the visual token pruning problem.

\begin{definition}[Visual Token Pruning]
Let $\mathcal{V} = \{v_1, v_2, \dots, v_N\}$ denote the set of $N$ visual tokens, where each token $v_i \in \mathbb{R}^d$ represents a $d$-dimensional feature vector.
Visual token pruning aims to select a subset $\mathcal{S} \subset \mathcal{V}$ with $|\mathcal{S}| = K < N$ tokens, where $K$ is a pre-defined budget constraint.
\end{definition}

To decouple the combining strategy from any specific importance estimator, we pre-define an importance vector $\mathbf{w}$ representing the weight of each token, regardless of how $\mathbf{w}$ is calculated.
Based on this, we define the retention metric:

\begin{definition}[Importance Retention Ratio]
The importance retention ratio of a subset $\mathcal{S}$ is defined as the normalized sum of retained scores:
$$ \mathcal{I}(\mathcal{S}) = \frac{\sum_{v_k \in \mathcal{S}} w_k}{\sum_{v_i \in \mathcal{V}} w_i} $$
This metric quantifies the proportion of total information retained by the subset, ranging from 0 to 1.
\end{definition}

In contrast to importance, which focuses on individual token utility, we characterize the spatial distribution of the selected subset using the Hopkins Statistic \cite{Hopkins1954ANM}, a measure that quantifies the degree of clustering in a dataset.
A high Hopkins value indicates strong clustering, meaning that selected tokens concentrate in specific semantic regions and thus exhibit high redundancy.

\begin{definition}[Diversity Metric via Hopkins Statistic]
Let $\mathcal{S}$ denote the selected token subset with $|\mathcal{S}| = m$.
We construct a reference set $\mathcal{R}$ by randomly sampling $m$ points from the same feature space as $\mathcal{S}$. 
Let $d(x, \mathcal{Y})$ denote the cosine distance from point $x$ to its nearest neighbor in set $\mathcal{Y}$.
The Hopkins Statistic is defined as:
\begin{equation*}
    H(\mathcal{S}) = \frac{\sum_{r \in \mathcal{R}} d(r, \mathcal{S})}{\sum_{r \in \mathcal{R}} d(r, \mathcal{S}) + \sum_{v \in \mathcal{S}} d(v, \mathcal{S} \setminus \{v\})}
\end{equation*}
In this formulation, $\mathcal{S} \setminus \{v\}$ denotes the set difference, representing the subset $\mathcal{S}$ excluding the specific token $v$ to ensure the distance is calculated against its nearest neighbor.
\end{definition}

Intuitively, $H(\mathcal{S}) \to 1$ indicates high redundancy due to significant clustering, while $H(\mathcal{S}) \to 0$ signifies a regularly spaced distribution with maximal semantic diversity.

\subsection{Simulation on Real Token Manifolds}

To identify the optimal strategy for harmonizing importance and diversity, we conduct a systematic analysis to explore their interaction.
Specifically, we employ real visual tokens extracted from the Vision Transformer of the Qwen2.5-VL-7B-Instruct model as feature vectors.
Real features are essential as they preserve complex manifold structures—such as semantic clustering and sparsity—that synthetic data typically fails to capture.

For token importance, we adopt a randomized approach where the score for each token is sampled independently from a uniform distribution $\mathcal{U}(0, 1)$.
This setup decouples the evaluation of selection strategies from the bias of any specific pre-trained importance scorer.

We evaluate five representative strategies that make different trade-offs between importance and diversity:
\begin{itemize}
    \item \textbf{Greedy Importance:} Selects tokens with the highest importance scores, ignoring diversity.
    
    \item \textbf{Greedy Diversity:} Iteratively selects the token that maximizes distance to the current subset via Farthest Point Sampling \cite{resende2010grasp}, prioritizing diversity over importance.
    
    \item \textbf{Naive Hybrid:} A two-stage approach that first selects top-$k$ tokens by importance, then applies Farthest Point Sampling within this subset.
    
    \item \textbf{Determinantal Point Processes (DPP):} Models diversity probabilistically via the determinant of a kernel matrix \cite{macchi1975coincidence}.
    
    \item \textbf{Maximal Marginal Relevance (MMR):} A joint optimization framework that explicitly balances importance and redundancy. We provide the detailed formulation of this mechanism in Section 3.3.
\end{itemize}

\subsection{The Maximal Marginal Relevance (MMR) Mechanism}

Maximal Marginal Relevance (MMR) \cite{Carbonell1998TheUO} provides a framework for this joint optimization.
Initially proposed for information retrieval, the core idea of MMR is that an ideal result set should balance two criteria: high relevance to the query and low redundancy among selected items.

Adapting this principle to visual token pruning, the algorithm iteratively selects the token $v^*$ from the candidate set $\mathcal{V} \setminus \mathcal{S}$ that maximizes the following objective:
\begin{equation*}
\begin{aligned}
    v^* = & \arg \max_{v_i \in \mathcal{V} \setminus \mathcal{S}} \big[ \lambda \cdot \text{Imp}(v_i) \\
    & - (1 - \lambda) \cdot \max_{v_j \in \mathcal{S}} \text{Sim}(v_i, v_j) \big]
\end{aligned}
\end{equation*}
where $\mathcal{V}$ represents the set of all visual tokens, $\mathcal{S}$ denotes the currently selected subset, $\text{Imp}(\cdot)$ represents the normalized importance score, $\text{Sim}(\cdot, \cdot)$ measures the pairwise similarity between tokens, and $\lambda$ is a hyperparameter balancing the two terms.

By subtracting the maximum similarity between the candidate and the current subset $\mathcal{S}$, the algorithm explicitly penalizes tokens that are semantically close to any already selected token, while prioritizing important tokens.

\subsection{Comparative Analysis against Heuristic Baselines}

We conducted the simulation on 200 randomly sampled images from the MMBench dataset \cite{MMBench} to systematically evaluate the efficacy of the proposed strategies.

\begin{figure}[h]
    \centering
    \includegraphics[width=0.98\linewidth]{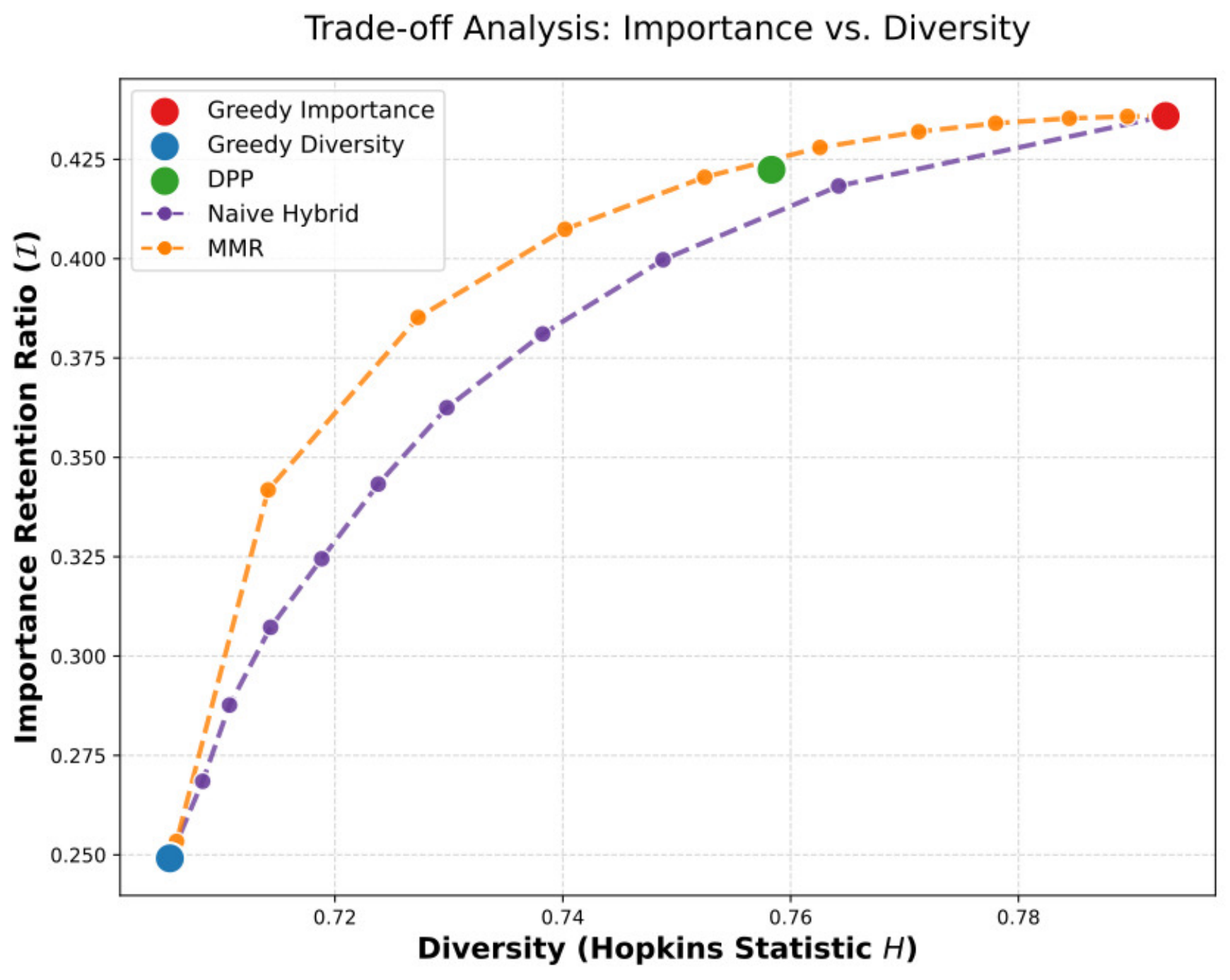} 
        \caption{
        \textbf{Pareto Frontier Analysis.}
        We visualize the trade-off between the \textbf{Hopkins Statistic ($H$)} and the \textbf{Importance Retention Ratio ($\mathcal{I}$)}.
        The ideal pruning strategy should approach the \textbf{top-left corner}, achieving a high Importance Retention Ratio ($\mathcal{I} \to 1$) while minimizing the Hopkins Statistic ($H \to 0$).
        The MMR mechanism (Orange) constructs a superior Pareto frontier that strictly dominates the Naive Hybrid strategy (Purple) and envelopes the DPP solution (Green).
    }
    \label{fig:simulation_tradeoff}
\end{figure}

Figure \ref{fig:simulation_tradeoff} illustrates the trade-off between importance retention and diversity for each strategy.
The theoretical optimum resides in the top-left corner, corresponding to subsets that maximize $\mathcal{I}$ while minimizing $H$, thereby maximizing diversity.
As illustrated, the single-objective baselines occupy the sub-optimal extremes: Greedy Importance (Red node) achieves maximum $\mathcal{I}$ at the cost of a high Hopkins Statistic ($H \approx 1$), whereas Greedy Diversity (Blue node) minimizes $H$ but suffers from a low Importance Retention Ratio.

Crucially, the trajectory generated by MMR (Orange curve) forms a superior Pareto Frontier.
It strictly dominates the Naive Hybrid strategy (Purple curve), maintaining a higher $\mathcal{I}$ for any given level of $H$, confirming the efficacy of our joint optimization framework.
Furthermore, it effectively envelopes the DPP solution (Green node), demonstrating that our joint optimization framework provides the most robust mechanism for harmonizing these conflicting objectives.

\section{Harmonizing Importance and Diversity via MMR}

\subsection{Token Importance Estimation}
\label{subsec:importance}

The analysis in Section \ref{sec:analysis} has demonstrated that the MMR mechanism effectively harmonizes diversity and importance.
However, applying this framework in practice requires a computable importance metric.

To this end, we adopt the importance estimation mechanism of VisionSelector \cite{zhu2025visionselectorendtoendlearnablevisual}, which currently represents the state-of-the-art among importance-based pruning approaches.
Specifically, this method employs a trainable estimation module coupled with a differentiable selection mechanism, \textit{DiffTopK}, to learn token importance through end-to-end training.
To maintain consistency with the training phase, we utilize the output of the \textit{DiffTopK} mechanism as our raw importance scores, denoted as $\mathbf{w}$.

However, since MMR involves a direct subtraction between importance and similarity, both metrics must have comparable scales to prevent one from dominating the selection process.
We therefore apply min-max normalization to the raw importance vector $\mathbf{w}$ to define the normalized importance metric:
\begin{equation*}
    \text{Imp}(v_i) = \frac{w_i - \min(\mathbf{w})}{\max(\mathbf{w}) - \min(\mathbf{w}) + \epsilon}
\end{equation*}
where $\epsilon$ is a small constant for numerical stability.
This procedure maps importance scores to the interval $[0, 1]$, ensuring they are commensurate with the similarity constraint.

\subsection{Quantifying Collective Redundancy}
\label{subsec:redundancy}

In addition to importance, the MMR framework requires a metric to quantify semantic redundancy. In the latent feature space of MLLMs, tokens representing similar visual concepts tend to cluster together. Thus, we define the pairwise similarity between a candidate token $v_i$ and a reference token $v_j$ using cosine similarity:
\begin{equation*}
    \text{Sim}(v_i, v_j) = \frac{v_i^\top v_j}{\|v_i\| \|v_j\|}
\end{equation*}
where $\|\cdot\|$ denotes the Euclidean norm. This metric enables the algorithm to identify tokens that are semantically similar to those already selected.

\begin{table*}[t]
\resizebox{\textwidth}{!}{
\begin{tabular}{lcccccccccccc}
\toprule
\multirow{2}{*}{\textbf{Method}} & \textbf{AI2D} & \textbf{ChartQA} & \textbf{DocVQA} & \textbf{MMB\textsuperscript{CN}} & \textbf{MMB} & \textbf{MME} & \textbf{MMStar} & \textbf{OCRBench} & \textbf{POPE} & \textbf{SQA} & \textbf{VQA\textsuperscript{Text}} & \multirow{2}{*}{\textbf{Avg}} \\
 & \textit{EM} & \textit{Relaxed} & \textit{Anls} & \textit{Score} & \textit{Score} & \textit{Score} & \textit{Avg} & \textit{Acc} & \textit{Acc} & \textit{EM} & \textit{EM} & \\
\midrule
Baseline & 82.48 & 83.68 & 94.90 & 80.41 & 83.08 & 1702 & 61.88 & 85.30 & 87.80 & 88.45 & 82.74 & 100.0\% \\
\midrule
\multicolumn{13}{c}{\cellcolor{gray!15} \textbf{Retain 25\% Tokens (75\% Compression Ratio) } } \\
\multicolumn{13}{l}{\textit{Importance-based methods}} \\
FastV & 75.68 & 68.20 & 81.20 & 73.20 & 76.12 & 1636 & 51.08 & 43.00 & 85.20 & 83.49 & 80.06 & 87.16\% \\
VisionZip & 77.40 & 67.20 & 71.48 & 76.12 & 78.78 & 1637 & 54.86 & 46.50 & 85.76 & 83.99 & 76.21 & 87.55\% \\
HiPrune & 77.49 & 68.60 & 73.52 & 76.03 & 78.09 & 1619 & 54.43 & 47.10 & 86.02 & 84.18 & 76.43 & 87.80\% \\
VisionSelector & \underline{79.60} & \underline{72.00} & \textbf{93.24} & 75.86 & 78.78 & \underline{1688} & 55.78 & \underline{72.50} & 86.74 & 85.08 & \underline{80.39} & \underline{94.22\%} \\
\multicolumn{13}{l}{\textit{Diversity-based methods}} \\
DivPrune & 77.98 & 62.00 & 85.32 & 75.77 & 77.84 & 1650 & 52.97 & 58.40 & 85.88 & 83.94 & 75.88 & 89.26\% \\
DART & 74.35 & 60.80 & 78.90 & 73.88 & 76.72 & 1625 & 52.90 & 46.00 & 84.34 & 84.33 & 71.68 & 85.74\% \\
\multicolumn{13}{l}{\textit{Hybrid strategies}} \\
VisPruner & 77.62 & 68.04 & 77.39 & 75.69 & 78.87 & 1657 & 54.01 & 48.70 & 85.68 & 84.18 & 75.17 & 88.31\% \\
SCOPE & 78.92 & 71.20 & 85.40 & \textbf{77.75} & \underline{79.38} & 1684 & \textbf{56.86} & 61.70 & \underline{86.78} & \underline{85.23} & 79.66 & 92.51\% \\
\textbf{IDPruner} & \textbf{80.51} & \textbf{74.32} & \underline{93.16} & \underline{76.63} & \textbf{79.73} & \textbf{1695} & \underline{56.49} & \textbf{74.00} & \textbf{87.06} & \textbf{85.52} & \textbf{80.83} & \textbf{95.18\%} \\
\midrule
\multicolumn{13}{c}{\cellcolor{gray!15} \textbf{Retain 10\% Tokens (90\% Compression Ratio) } } \\
\multicolumn{13}{l}{\textit{Importance-based methods}} \\
FastV & 67.23 & 39.48 & 51.90 & 53.26 & 55.58 & 1332 & 38.02 & 24.10 & 76.31 & 79.28 & 72.59 & 68.07\% \\
VisionZip & 70.60 & 41.56 & 37.94 & 66.67 & 71.05 & 1462 & 45.19 & 23.40 & 81.06 & \textbf{83.24} & 61.06 & 71.84\% \\
HiPrune & 69.82 & 43.96 & 39.89 & 67.44 & 70.88 & 1438 & 45.04 & 23.70 & 80.70 & 82.65 & 62.51 & 72.22\% \\
VisionSelector & \underline{74.81} & \textbf{62.68} & \textbf{87.00} & 68.99 & 71.65 & 1569 & 46.93 & \textbf{55.50} & 82.69 & 81.95 & \textbf{74.52} & \underline{85.39\%} \\
\multicolumn{13}{l}{\textit{Diversity-based methods}} \\
DivPrune & 70.11 & 41.36 & 66.20 & 69.42 & 72.16 & 1529 & 44.46 & 31.80 & 81.91 & 80.96 & 62.72 & 76.09\% \\
DART & 67.88 & 34.84 & 49.86 & 63.92 & 67.35 & 1451 & 42.93 & 24.30 & 79.70 & 80.96 & 54.06 & 69.80\% \\
\multicolumn{13}{l}{\textit{Hybrid strategies}} \\
VisPruner & 69.88 & 42.68 & 50.85 & 66.84 & 70.96 & 1442 & 44.14 & 24.40 & 81.03 & 81.11 & 59.66 & 72.60\% \\
SCOPE & 71.63 & 50.04 & 56.45 & \underline{71.22} & \textbf{75.43} & \underline{1608} & \textbf{48.74} & 34.10 & \underline{84.10} & 82.25 & 70.61 & 79.35\% \\
\textbf{IDPruner} & \textbf{75.16} & \underline{62.48} & \underline{85.98} & \textbf{71.65} & \underline{74.66} & \textbf{1618} & \underline{47.48} & \underline{53.90} & \textbf{85.43} & \underline{82.80} & \underline{74.43} & \textbf{86.47\%} \\
\bottomrule
\end{tabular}
}
\centering
\caption{Comparison results on comprehensive Image-Language benchmarks on \textbf{Qwen-2.5-7B-Instruct}.}
\label{tab:results_qwen2.5_7b}
\end{table*}

\subsection{IDPruner: An MMR-based Selection Strategy}
\label{subsec:id_pruner}

Building upon the normalized importance and semantic similarity metrics defined above, we formally present the \textbf{I}mportance and \textbf{D}iversity Pruner (\textbf{IDPruner}).
This method harmonizes the two conflicting objectives within the MMR framework to iteratively construct the optimal subset.
At each step $t$, IDPruner selects the token $v^*$ from the remaining candidates $\mathcal{V} \setminus \mathcal{S}_{t-1}$ by maximizing the following objective:
\begin{equation*}
v^* = \arg \max_{v_i \in \mathcal{V} \setminus \mathcal{S}_{t-1}} [ \lambda \cdot \text{Imp}(v_i) - (1 - \lambda) \cdot m_i ]
\end{equation*}
where $m_i = \max_{v_j \in \mathcal{S}_{t-1}} \text{Sim}(v_i, v_j)$ represents the maximum similarity between the candidate $v_i$ and any token in the currently selected set, and $\lambda \in [0, 1]$ is the hyperparameter balancing importance and diversity.

\begin{algorithm}[h]
\caption{IDPruner}
\label{alg:id_pruner}
\begin{algorithmic}[1]
\REQUIRE Tokens $\mathcal{V}$, Raw Importance Scores $\mathbf{w}$, Budget $K$, Hyperparameter $\lambda$
\ENSURE Pruned subset $\mathcal{S}$
\STATE $\text{Imp} \leftarrow (\mathbf{w} - \min \mathbf{w}) / (\max \mathbf{w} - \min \mathbf{w} + \epsilon)$
\STATE $\mathcal{S} \leftarrow \emptyset, \mathbf{m} \leftarrow \text{fill}(N, -1.0)$
\FOR{$t = 1$ to $K$}
    \IF{$t = 1$}
        \STATE $v^* \leftarrow \arg \max_{v_i \in \mathcal{V}} \text{Imp}(v_i)$
    \ELSE
        \STATE $v^* \leftarrow \arg \max_{v_i \notin \mathcal{S}} [ \lambda \text{Imp}(v_i) - (1 - \lambda) m_i ]$
    \ENDIF
    \STATE $\mathcal{S} \leftarrow \mathcal{S} \cup \{v^*\}$
    \STATE $\mathbf{m} \leftarrow \max(\mathbf{m}, \text{Sim}(\mathcal{V}, v^*))$
\ENDFOR
\RETURN $\mathcal{S}$
\end{algorithmic}
\end{algorithm}

To minimize computational overhead, we adopt an efficient updating strategy.
Instead of recomputing the similarity scores for all pairs at every step, we maintain a vector $\mathbf{m} \in \mathbb{R}^N$ that tracks the maximum similarity for each candidate.
After selecting $v^*$, we simply update this vector: $m_i \leftarrow \max(m_i, \text{Sim}(v_i, v^*))$.
This implementation reduces the computational complexity from $O(K^2 N)$ to $O(KN)$, rendering the overhead negligible relative to the model's forward pass.
The complete procedure is summarized in Algorithm \ref{alg:id_pruner}.

\begin{table*}[t]
\label{tab:results_llava_1.5_7b}
\resizebox{\textwidth}{!}{
\begin{tabular}{lcccccccccccc}
\toprule
\multirow{2}{*}{\textbf{Method}} & \textbf{AI2D} & \textbf{ChartQA} & \textbf{DocVQA} & \textbf{MMB\textsuperscript{CN}} & \textbf{MMB} & \textbf{MME} & \textbf{MMStar} & \textbf{OCRBench} & \textbf{POPE} & \textbf{SQA} & \textbf{VQA\textsuperscript{Text}} & \multirow{2}{*}{\textbf{Avg}} \\
 & \textit{EM} & \textit{Relaxed} & \textit{Anls} & \textit{Score} & \textit{Score} & \textit{Score} & \textit{Avg} & \textit{Acc} & \textit{Acc} & \textit{EM} & \textit{EM} & \\
\midrule
Baseline & 52.78 & 18.12 & 24.09 & 50.17 & 62.20 & 1463 & 32.74 & 19.80 & 85.86 & 66.19 & 47.78 & 100.0\% \\
\midrule
\multicolumn{13}{c}{\cellcolor{gray!15} \textbf{Retain 128 Tokens (77\% Compression Ratio) } } \\
\multicolumn{13}{l}{\textit{Importance-based methods}} \\
FastV & 50.58 & 13.84 & 12.69 & 46.13 & 57.90 & 1213 & 30.89 & 12.10 & 72.04 & 65.44 & 31.76 & 81.59\% \\
VisionZip & 50.81 & 16.80 & 19.93 & 48.88 & 60.05 & 1374 & 32.52 & 18.50 & 82.28 & 66.34 & 45.40 & 94.86\% \\
HiPrune & \underline{51.98} & 17.00 & 20.79 & \textbf{49.74} & 60.14 & \underline{1386} & 32.20 & 18.50 & 82.14 & 66.19 & 45.54 & 95.63\% \\
VisionSelector & 50.74 & 16.20 & 20.83 & \underline{49.23} & 60.31 & 1379 & \textbf{34.22} & 18.40 & 83.06 & 66.39 & 45.29 & 95.51\% \\
\multicolumn{13}{l}{\textit{Diversity-based methods}} \\
DivPrune & 51.65 & 16.36 & 18.58 & 46.13 & 57.99 & 1354 & 32.79 & 17.70 & \textbf{85.16} & \underline{66.63} & 43.75 & 93.09\% \\
DART & \textbf{52.82} & 15.32 & 15.02 & 44.50 & 56.53 & 1309 & 30.55 & 14.00 & 77.08 & 66.04 & 34.90 & 85.69\% \\
\multicolumn{13}{l}{\textit{Hybrid strategies}} \\
VisPruner & 51.68 & 16.56 & 20.05 & \textbf{49.74} & \textbf{60.57} & 1382 & 32.47 & 18.20 & 83.57 & \underline{66.63} & 46.21 & 95.39\% \\
SCOPE & 51.30 & \underline{17.20} & \textbf{21.56} & 49.14 & 60.22 & 1374 & 32.68 & \textbf{19.30} & 84.41 & 66.58 & \textbf{46.84} & \underline{96.77\%} \\
\textbf{IDPruner} & 51.55 & \textbf{17.40} & \underline{21.55} & \underline{49.23} & \underline{60.40} & \textbf{1428} & \underline{33.44} & \underline{18.80} & \underline{84.69} & \textbf{66.88} & \underline{46.39} & \textbf{97.26\%} \\
\midrule
\multicolumn{13}{c}{\cellcolor{gray!15} \textbf{Retain 64 Tokens (88\% Compression Ratio) } } \\
\multicolumn{13}{l}{\textit{Importance-based methods}} \\
FastV & 49.42 & 12.24 & 9.74 & 40.38 & 47.08 & 964 & 27.78 & 4.00 & 61.38 & 63.71 & 17.69 & 66.68\% \\
VisionZip & 51.20 & \underline{15.76} & 15.75 & 46.31 & 56.70 & 1289 & 31.40 & 16.40 & 78.18 & 66.44 & 42.75 & 89.14\% \\
HiPrune & 51.20 & 15.68 & 16.18 & 46.13 & 57.30 & 1257 & 31.15 & \underline{17.40} & 78.17 & 66.48 & 43.00 & 89.56\% \\
VisionSelector & 50.65 & 15.12 & 17.29 & 46.65 & \textbf{58.33} & 1310 & \textbf{32.58} & 17.00 & 79.94 & 65.84 & 42.98 & 90.49\% \\
\multicolumn{13}{l}{\textit{Diversity-based methods}} \\
DivPrune & 50.00 & 15.12 & 15.69 & 44.16 & 54.64 & 1271 & 31.68 & 15.70 & \textbf{84.18} & 66.44 & 40.83 & 87.82\% \\
DART & \textbf{51.55} & 15.20 & 12.35 & 40.29 & 53.26 & 1195 & 28.50 & 12.40 & 71.03 & \textbf{66.83} & 30.98 & 79.88\% \\
\multicolumn{13}{l}{\textit{Hybrid strategies}} \\
VisPruner & \underline{51.33} & 15.28 & 16.29 & 44.76 & 57.39 & 1315 & 31.02 & 17.00 & 81.20 & \underline{66.73} & 43.45 & 89.77\% \\
SCOPE & 50.71 & 15.56 & \underline{17.84} & \underline{46.82} & \underline{58.16} & \underline{1320} & \underline{32.18} & \textbf{17.70} & 83.07 & 66.63 & \underline{44.33} & \underline{91.90\%} \\
\textbf{IDPruner} & 50.55 & \textbf{16.32} & \textbf{18.59} & \textbf{46.99} & 57.65 & \textbf{1329} & 31.49 & 17.30 & \underline{83.83} & 66.68 & \textbf{44.73} & \textbf{92.34\%} \\
\midrule
\multicolumn{13}{c}{\cellcolor{gray!15} \textbf{Retain 32 Tokens (94\% Compression Ratio) } } \\
\multicolumn{13}{l}{\textit{Importance-based methods}} \\
FastV & 49.03 & 11.96 & 8.38 & 30.24 & 38.23 & 844 & 27.15 & 3.00 & 57.50 & 64.20 & 10.98 & 59.83\% \\
VisionZip & 51.17 & 14.04 & 13.10 & 39.35 & 50.43 & 1133 & 29.90 & 13.90 & 72.42 & 66.98 & 38.15 & 81.15\% \\
HiPrune & \underline{51.42} & 14.32 & 13.22 & 39.95 & 51.98 & 1149 & 29.28 & 14.20 & 73.26 & 66.78 & 38.29 & 81.87\% \\
VisionSelector & 50.06 & 13.52 & 14.00 & \underline{43.81} & 54.90 & 1194 & \textbf{32.07} & 14.80 & 75.04 & 65.49 & 38.94 & 84.12\% \\
\multicolumn{13}{l}{\textit{Diversity-based methods}} \\
DivPrune & 50.74 & 14.20 & 12.25 & 37.89 & 51.72 & 1149 & 28.68 & 13.80 & \underline{80.47} & 65.69 & 36.21 & 80.78\% \\
DART & 51.10 & 13.32 & 10.79 & 33.51 & 47.08 & 1069 & 29.17 & 10.70 & 63.86 & \textbf{67.58} & 25.35 & 73.03\% \\
\multicolumn{13}{l}{\textit{Hybrid strategies}} \\
VisPruner & \textbf{51.59} & 14.28 & 12.95 & 37.97 & 50.95 & 1165 & 29.40 & 14.00 & 75.14 & \underline{67.23} & 37.75 & 81.47\% \\
SCOPE & 51.17 & \textbf{15.00} & \underline{14.91} & 43.64 & \textbf{55.58} & \underline{1239} & \underline{31.36} & \underline{15.10} & 79.50 & 66.83 & \underline{40.14} & \underline{86.57\%} \\
\textbf{IDPruner} & 51.23 & \underline{14.80} & \textbf{15.11} & \textbf{44.76} & \underline{55.50} & \textbf{1254} & 30.58 & \textbf{15.90} & \textbf{81.43} & 66.73 & \textbf{41.46} & \textbf{87.43\%} \\
\bottomrule
\end{tabular}
}
\centering
\caption{Comparison results on comprehensive Image-Language benchmarks on \textbf{LLaVA-1.5-7B}.}

\label{tab:results_llava_1.5_7b}
\end{table*}

\section{Experiments}

\subsection{Experimental Setup}

\noindent \textbf{Model Architectures.}
We conduct our main experiments on widely adopted MLLMs, including Qwen2.5-VL-7B-Instruct \cite{bai2025qwen2} and LLaVA-1.5-7B. \cite{liu2023llava}.

\noindent \textbf{Evaluation benchmarks.}
We conduct comprehensive evaluations on image and video understanding tasks.
For image-language understanding, we employ 10 widely-used datasets: MME \cite{Fu2023MME}, MMBench \cite{MMBench}, MMStar \cite{chen2024mmstar}, POPE \cite{li2023pope}, ScienceQA \cite{lu2022sqa}, AI2D \cite{Kembhavi2016ai2d}, TextVQA \cite{singh2019textvqa}, ChartQA \cite{masry2022chartqa}, DocVQA \cite{mathew2020docvqa}, and OCRBench \cite{Liu_2024ocrbench}.
For video-language understanding, we include 3 benchmarks: Vinoground \cite{zhang2024vinoground}, VideoMME \cite{fu2025video}, and SEED-Bench \cite{li2024seed}.
To ensure fair comparison and reproducibility, we utilize the LMMs-Eval framework \citep{zhang2024lmmsevalrealitycheckevaluation}, strictly following the default settings and metrics for each task.

\noindent \textbf{Comparison methods.}
We compare IDPruner with representative state-of-the-art approaches across different paradigms, including importance-based methods like FastV \citep{chen2024image}, VisionZip \citep{yang2025visionzip}, HiPrune \citep{liu2025hiprune}, and VisionSelector \citep{zhu2025visionselectorendtoendlearnablevisual}, diversity-based methods like DivPrune \citep{Alvar2025DivPruneDV} and DART \citep{Wen2025StopLF}, as well as hybrid strategies that combine multiple criteria, such as VisPruner \citep{Zhang2024BeyondTA}, and SCOPE \citep{Deng2025SCOPESO}.

\noindent \textbf{Implementation Details.}
Unless otherwise specified, the hyperparameter $\lambda$ of IDPruner, which balances importance and diversity, is set to 0.5.

\subsection{Main Results}

\noindent \textbf{Results on Qwen2.5-VL-7B-Instruct.}
We evaluate our method on Qwen2.5-VL-7B-Instruct under 25\% and 10\% token retention settings.
As shown in Table \ref{tab:results_qwen2.5_7b}, IDPruner achieves state-of-the-art average scores of 95.18\% and 86.47\%, respectively.
Compared to existing strategies, our method achieves a better balance between keeping fine details and maintaining global context.
Specifically, for tasks requiring fine details, such as OCRBench, our method ranks among the top two, while also maintaining global information to surpass VisionSelector on hallucination benchmarks, including POPE.
Consequently, on benchmarks such as MME and AI2D, which require both overall understanding and detailed capture, IDPruner demonstrates a clear lead over other methods.

\noindent \textbf{Results on LLaVA-1.5-7B.}
We extend our experiments to the LLaVA-1.5-7B model, which operates with a fixed resolution of 576 visual tokens per image.
Accordingly, we evaluate performance under three distinct retention settings: 128, 64, and the extreme 32 tokens.
As shown in Table \ref{tab:results_llava_1.5_7b}, IDPruner consistently achieves state-of-the-art results across all pruning ratios.
While VisionSelector is surpassed by the hybrid method SCOPE on this architecture, IDPruner maintains its lead, achieving an average score of 87.43\% even with only 32 tokens, demonstrating its robustness across diverse architectures.

In summary, our method exhibits remarkable performance consistency across a diverse range of architectures.
Notably, strong baselines exhibit architecture-specific vulnerabilities; for instance, VisionSelector underperforms on LLaVA-1.5, whereas SCOPE loses competitiveness on the advanced LLaVA-OneVision-1.5, as detailed in Appendix~\ref{sec:appendix_additional_results}.
In contrast, IDPruner maintains exceptional robustness.
It consistently achieves state-of-the-art results across all evaluated models, validating the universality of our framework in harmonizing token importance and diversity.

\begin{table}[h] 
\centering
\resizebox{\linewidth}{!}{
\begin{tabular}{lcccc}
\toprule
\multirow{2}{*}{\textbf{Method}} & \textbf{Vinoground} & \textbf{VideoMME} & \textbf{SEED-Bench} & \multirow{2}{*}{\textbf{Avg}} \\
 & \textit{Group} & \textit{Perception} & \textit{All} & \\
\midrule
Baseline & 20.20 & 61.33 & 74.12 & 100.0\% \\
\midrule
FastV & 12.80 & 59.44 & 69.22 & 84.56\% \\
VisionZip & 12.80 & 59.67 & 72.11 & 85.98\% \\
HiPrune & 11.80 & \underline{59.93} & 71.97 & 84.41\% \\
VisionSelector & 10.80 & 59.19 & 70.75 & 81.81\% \\
DivPrune & \textbf{14.00} & 58.00 & 72.11 & \underline{87.06\%} \\
DART & 12.60 & 59.52 & 71.27 & 85.19\% \\
VisPruner & 11.40 & 59.44 & 71.89 & 83.45\% \\
SCOPE & 12.80 & \textbf{60.00} & \underline{72.63} & 86.40\% \\
\textbf{IDPruner} & \underline{13.40} & 59.48 & \textbf{72.68} & \textbf{87.13\%} \\
\bottomrule
\end{tabular}
}
\caption{Comparison results on Video-Language benchmarks on \textbf{Qwen2.5-VL-7B-Instruct} with \textbf{25\% token retention}.}
\label{tab:video_results} 
\end{table}

\subsection{IDPruner for Video Understanding}

Beyond static image benchmarks, we extend IDPruner to video understanding tasks, evaluating its performance on Vinoground, VideoMME, and SEED-Bench at a 75\% pruning ratio.
As shown in Table \ref{tab:video_results}, purely importance-based methods exhibit significant performance degradation. This is primarily due to their inability to handle the high temporal redundancy in videos.
In contrast, diversity-based methods maintain strong performance with an average score of 87.06\%.
Notably, IDPruner achieves the best average performance of 87.13\% by jointly considering both the preservation of important details and the reduction of temporal redundancy.

\begin{table}[h]
\centering
\setlength{\tabcolsep}{16pt}      

\resizebox{0.9\linewidth}{!}{
\begin{tabular}{l c c c}
\toprule
\textbf{Method} & \textbf{FA} & \textbf{Prefill}(ms) & \textbf{E2E Latency}(ms) \\
\midrule
VisPruner & $\times$ & 1459.95 & 1600.81 \\
SCOPE & $\times$ & 1677.81 & 1818.40 \\
\textbf{IDPruner} & \checkmark & \textbf{1337.76} & \textbf{1478.32} \\
\bottomrule
\end{tabular}
}
\caption{Efficiency analysis on Vinoground on \textbf{Qwen2.5-VL-7B-Instruct} with \textbf{25\% token retention}. \textbf{FA}: FlashAttention compatibility.}
\label{tab:efficiency_hybrid}
\end{table}

\subsection{Efficiency and Practicality}

We compare the efficiency of hybrid pruning strategies on Qwen2.5-VL-7B using the Vinoground benchmark at a 75\% pruning ratio.
As shown in Table \ref{tab:efficiency_hybrid}, IDPruner achieves the best efficiency among hybrid strategies, due to its lightweight diversity calculation and being attention-map-free.
This design ensures full compatibility with FlashAttention, yielding the lowest prefill time of 1337.76 ms and an end-to-end latency of 1478.32 ms.

\section{Conclusion}

Recent progress in visual token pruning shows that hybrid strategies are surpassing methods that rely only on importance or diversity, becoming the new standard in this field.
However, there is a lack of systematic analysis on how to effectively harmonize these two objectives.
In this study, we provide a framework to analyze this trade-off and demonstrate that the Maximal Marginal Relevance (MMR) mechanism is an effective strategy to achieve an optimal balance.
Based on this insight, we propose IDPruner, a method that explicitly balances token importance and semantic redundancy.
Extensive evaluations show that our method achieves state-of-the-art performance and remains robust across different model architectures.
We believe this work offers a solid foundation for systematically balancing importance and diversity, enabling more efficient MLLMs.


\bibliography{latex/custom}

\appendix
\newpage

\appendix

\clearpage

\begin{table*}[p]
\centering
\resizebox{\textwidth}{!}{
\begin{tabular}{lcccccccccccc}
\toprule
\multirow{2}{*}{\textbf{Method}} & \textbf{AI2D} & \textbf{ChartQA} & \textbf{DocVQA} & \textbf{MMB\textsuperscript{CN}} & \textbf{MMB} & \textbf{MME} & \textbf{MMStar} & \textbf{OCRBench} & \textbf{POPE} & \textbf{SQA} & \textbf{VQA\textsuperscript{Text}} & \multirow{2}{*}{\textbf{Avg}} \\
 & \textit{EM} & \textit{Relaxed} & \textit{Anls} & \textit{Score} & \textit{Score} & \textit{Score} & \textit{Avg} & \textit{Acc} & \textit{Acc} & \textit{EM} & \textit{EM} & \\
\midrule
Baseline & 79.11 & 83.56 & 92.48 & 73.28 & 77.32 & 1517 & 56.05 & 80.10 & 87.41 & 80.81 & 78.79 & 100.0\% \\
\midrule
\multicolumn{13}{c}{\cellcolor{gray!15} \textbf{Retain 25\% Tokens (75\% Compression Ratio) } } \\
\multicolumn{13}{l}{\textit{Importance-based methods}} \\
FastV & 72.70 & 70.04 & 75.98 & 63.40 & 66.92 & 1437 & 47.39 & 36.60 & 86.42 & 79.33 & 73.51 & 86.02\% \\
VisionZip & 74.19 & 71.32 & 70.11 & 67.35 & 71.22 & 1452 & 49.37 & 42.50 & 85.51 & \underline{81.36} & 68.12 & 87.34\% \\
HiPrune & 73.83 & 72.76 & 72.10 & 67.27 & 72.34 & 1449 & 48.93 & 41.30 & 85.86 & 80.91 & 69.27 & 87.67\% \\
VisionSelector & 75.19 & 73.72 & \textbf{90.24} & \underline{68.81} & 72.59 & \textbf{1521} & \underline{49.97} & \underline{61.80} & 85.36 & 80.37 & \underline{76.86} & \underline{93.62\%} \\
\multicolumn{13}{l}{\textit{Diversity-based methods}} \\
DivPrune & 73.06 & 62.96 & 78.46 & 67.10 & 71.82 & 1459 & 48.38 & 51.40 & \textbf{86.81} & 80.22 & 68.91 & 88.15\% \\
DART & 71.08 & 65.20 & 79.72 & 65.38 & 71.05 & 1428 & 48.78 & 41.80 & 80.97 & 80.91 & 68.25 & 86.17\% \\
\multicolumn{13}{l}{\textit{Hybrid strategies}} \\
VisPruner & 74.29 & 68.20 & 72.52 & 67.35 & 70.88 & 1458 & 49.74 & 44.80 & 86.59 & \textbf{81.46} & 69.62 & 87.87\% \\
SCOPE & \underline{75.84} & \underline{74.00} & 82.40 & \underline{68.81} & \underline{72.94} & 1471 & \textbf{50.35} & 56.00 & \underline{86.62} & 80.96 & 74.04 & 91.98\% \\
\textbf{IDPruner} & \textbf{75.94} & \textbf{75.84} & \underline{90.00} & \textbf{69.42} & \textbf{73.80} & \underline{1505} & 49.49 & \textbf{64.90} & 86.26 & 80.42 & \textbf{76.90} & \textbf{94.42\%} \\
\midrule
\multicolumn{13}{c}{\cellcolor{gray!15} \textbf{Retain 10\% Tokens (90\% Compression Ratio) } } \\
\multicolumn{13}{l}{\textit{Importance-based methods}} \\
FastV & 65.87 & 29.72 & 36.89 & 48.37 & 51.98 & 1257 & 37.28 & 13.90 & 79.50 & 77.05 & 57.75 & 65.30\% \\
VisionZip & 67.65 & 51.60 & 37.88 & 59.62 & 63.06 & 1338 & 42.82 & 21.40 & 81.14 & 80.47 & 51.56 & 72.75\% \\
HiPrune & 67.75 & 53.20 & 41.15 & 59.45 & 63.14 & 1326 & 41.08 & 20.30 & 80.90 & \textbf{80.96} & 53.31 & 73.00\% \\
VisionSelector & \underline{70.50} & \textbf{65.92} & \textbf{79.94} & 59.97 & 64.69 & 1374 & 42.86 & \underline{45.20} & 82.66 & \underline{80.61} & \textbf{71.57} & \underline{84.42\%} \\
\multicolumn{13}{l}{\textit{Diversity-based methods}} \\
DivPrune & 67.71 & 43.12 & 58.03 & 61.25 & 65.12 & 1389 & 40.43 & 27.90 & 82.24 & 79.18 & 56.87 & 75.50\% \\
DART & 67.49 & 47.56 & 60.23 & 57.99 & 63.83 & 1299 & 42.18 & 23.40 & 74.20 & 78.63 & 58.02 & 74.09\% \\
\multicolumn{13}{l}{\textit{Hybrid strategies}} \\
VisPruner & 67.75 & 47.92 & 48.65 & 59.28 & 63.32 & 1305 & 41.51 & 22.50 & 78.74 & 79.77 & 54.95 & 73.19\% \\
SCOPE & 69.75 & 56.24 & 55.01 & \textbf{64.26} & \underline{67.18} & \underline{1390} & \textbf{44.35} & 30.80 & \underline{83.34} & 80.47 & 62.58 & 79.37\% \\
\textbf{IDPruner} & \textbf{71.79} & \underline{63.32} & \underline{79.38} & \underline{63.57} & \textbf{68.21} & \textbf{1438} & \underline{44.05} & \textbf{45.50} & \textbf{84.51} & 80.57 & \underline{70.02} & \textbf{85.71\%} \\
\bottomrule
\end{tabular}
}
\caption{Comparison results with different methods on \textbf{Qwen2.5-VL-3B-Instruct}.}
\label{tab:results_qwen2.5_3b}
\end{table*}

\begin{table*}[p]
\centering
\resizebox{\textwidth}{!}{
\begin{tabular}{lcccccccccccc}
\toprule
\multirow{2}{*}{\textbf{Method}} & \textbf{AI2D} & \textbf{ChartQA} & \textbf{DocVQA} & \textbf{MMB\textsuperscript{CN}} & \textbf{MMB} & \textbf{MME} & \textbf{MMStar} & \textbf{OCRBench} & \textbf{POPE} & \textbf{SQA} & \textbf{VQA\textsuperscript{Text}} & \multirow{2}{*}{\textbf{Avg}} \\
 & \textit{EM} & \textit{Relaxed} & \textit{Anls} & \textit{Score} & \textit{Score} & \textit{Score} & \textit{Avg} & \textit{Acc} & \textit{Acc} & \textit{EM} & \textit{EM} & \\
\midrule
Baseline & 84.20 & 86.40 & 97.87 & 78.52 & 85.31 & 1594 & 68.25 & 80.90 & 88.91 & 98.76 & 79.65 & 100.0\% \\
\midrule
\multicolumn{13}{c}{\cellcolor{gray!15} \textbf{Retain 25\% Tokens (75\% Compression Ratio) } } \\
\multicolumn{13}{l}{\textit{Importance-based methods}} \\
FastV & 74.48 & 39.16 & 56.53 & 70.36 & 77.66 & 1440 & 50.21 & 33.60 & 81.53 & 89.19 & 52.54 & 75.05\% \\
VisionZip & 69.95 & 27.48 & 23.69 & 65.89 & 75.26 & 1419 & 47.01 & 20.80 & 80.46 & 84.78 & 36.15 & 65.14\% \\
HiPrune & 70.27 & 19.76 & 21.01 & 63.57 & 72.51 & 1339 & 46.58 & 19.60 & 77.29 & 81.36 & 30.21 & 61.59\% \\
VisionSelector & 77.85 & \textbf{76.32} & \textbf{94.08} & \textbf{74.74} & 80.07 & \underline{1569} & \underline{57.42} & \underline{55.50} & 86.73 & \underline{94.70} & \textbf{77.53} & \underline{91.63\%} \\
\multicolumn{13}{l}{\textit{Diversity-based methods}} \\
DivPrune & \underline{78.85} & 65.52 & 81.53 & 73.20 & \underline{80.67} & 1533 & 56.68 & 51.50 & \textbf{88.14} & 91.42 & 75.00 & 88.12\% \\
DART & 77.82 & 69.00 & 84.77 & 72.25 & 79.81 & 1564 & 56.80 & 46.90 & 84.77 & 93.11 & 73.89 & 87.83\% \\
\multicolumn{13}{l}{\textit{Hybrid strategies}} \\
VisPruner & 75.45 & 45.88 & 61.00 & 69.93 & 76.55 & 1468 & 50.68 & 36.70 & 85.82 & 89.24 & 69.01 & 79.01\% \\
SCOPE & 78.21 & 54.84 & 75.68 & 70.27 & 78.18 & 1505 & 54.52 & 46.60 & 87.02 & 90.18 & 72.77 & 84.30\% \\
\textbf{IDPruner} & \textbf{79.18} & \underline{74.48} & \underline{91.82} & \underline{73.71} & \textbf{81.27} & \textbf{1588} & \textbf{57.97} & \textbf{57.80} & \underline{88.13} & \textbf{95.14} & \underline{77.50} & \textbf{92.00\%} \\
\midrule
\multicolumn{13}{c}{\cellcolor{gray!15} \textbf{Retain 10\% Tokens (90\% Compression Ratio) } } \\
\multicolumn{13}{l}{\textit{Importance-based methods}} \\
FastV & 70.95 & 22.72 & 29.77 & 62.46 & 69.07 & 1303 & 43.66 & 16.70 & 73.83 & 82.35 & 38.71 & 62.08\% \\
VisionZip & 69.14 & 19.64 & 13.20 & 58.16 & 65.38 & 1259 & 42.12 & 12.00 & 76.59 & 78.53 & 21.28 & 56.09\% \\
HiPrune & 68.39 & 16.24 & 12.86 & 56.01 & 63.83 & 1208 & 42.13 & 9.10 & 73.72 & 77.29 & 18.97 & 53.92\% \\
VisionSelector & 73.38 & \textbf{61.00} & \textbf{71.31} & \underline{68.04} & \underline{74.91} & 1466 & \textbf{50.74} & \underline{34.60} & 82.62 & \underline{88.05} & \underline{67.67} & \underline{80.11\%} \\
\multicolumn{13}{l}{\textit{Diversity-based methods}} \\
DivPrune & \underline{73.74} & 37.68 & 56.57 & 67.18 & 73.80 & \underline{1477} & 46.31 & 30.10 & \textbf{85.33} & 86.51 & 65.98 & 75.02\% \\
DART & 72.80 & 40.48 & 52.77 & 63.14 & 70.19 & 1378 & 47.42 & 27.30 & 75.79 & 84.88 & 61.58 & 71.65\% \\
\multicolumn{13}{l}{\textit{Hybrid strategies}} \\
VisPruner & 69.82 & 25.20 & 35.18 & 61.77 & 67.70 & 1375 & 43.24 & 18.20 & 78.33 & 84.23 & 54.53 & 65.46\% \\
SCOPE & 72.05 & 33.04 & 47.65 & 66.07 & 73.54 & 1471 & 48.25 & 25.70 & 84.11 & 85.97 & 61.02 & 72.35\% \\
\textbf{IDPruner} & \textbf{74.45} & \underline{57.92} & \underline{70.43} & \textbf{69.67} & \textbf{75.69} & \textbf{1511} & \underline{50.02} & \textbf{38.70} & \underline{85.02} & \textbf{88.70} & \textbf{72.37} & \textbf{81.55\%} \\
\bottomrule
\end{tabular}
}
\caption{Comparison results with different methods on \textbf{LLaVA-OneVision-1.5-8B-Instruct}.}
\label{tab:results_llava_onevision}
\end{table*}

\section{Additional Experimental Results}
\label{sec:appendix_additional_results}

\subsection{Results on Qwen2.5-VL-3B-Instruct}

To evaluate the scalability of our method on smaller language models, we conduct experiments on \textbf{Qwen2.5-VL-3B-Instruct}.
As shown in Table \ref{tab:results_qwen2.5_3b}, IDPruner consistently outperforms competitive baselines at both 25\% and 10\% token retention ratios.
Notably, when retaining 25\% of the tokens, our method achieves an average score of 94.42\%, effectively matching the unpruned baseline.
Even under the aggressive 10\% retention setting, IDPruner maintains a high average performance of 85.71\%, outperforming the second-best method (VisionSelector) by 1.29\%.

\subsection{Results on LLaVA-OneVision-1.5-8B-Instruct}

We further assess the cross-architecture generalization on \textbf{LLaVA-OneVision-1.5-8B-Instruct}, which integrates advanced visual encoding strategies.
As shown in Table \ref{tab:results_llava_onevision}, IDPruner achieves the best results among existing state-of-the-art methods.
Under the 25\% retention setting, our method achieves an average score of 92.00\%, outperforming the strongest baseline, VisionSelector, by 0.37\%.
In the more challenging 10\% retention scenario, IDPruner exhibits strong robustness, achieving an average score of 81.55\%.
It significantly outperforms purely importance-based methods such as VisionZip and HiPrune, which suffer from severe degradation due to the loss of global context.
Additionally, it surpasses the competitive VisionSelector by 1.44\%, confirming that harmonizing importance and diversity is particularly effective for advanced architectures.

\section{Ablation Study: Integration Strategies and Hyperparameters}
\label{sec:ablation_study}

We investigate the efficacy of different integration strategies and the impact of the hyperparameter $\lambda$, which controls the trade-off between token importance and diversity.
Using VisionSelector as the fixed base importance estimator, we compare our \textbf{IDPruner (MMR)} mechanism against two representative baselines: a determinantal point process based method (\textbf{DPP}) and a \textbf{Naive Hybrid} strategy that combines importance filtering with Farthest Point Sampling (FPS).
Table \ref{tab:ablation_lambda} summarizes the results on \textbf{Qwen2.5-VL-7B-Instruct} at a 25\% token retention ratio.

\begin{table*}[t]
\centering
\resizebox{\textwidth}{!}{
\begin{tabular}{lccccccccc}
\toprule
\multirow{2}{*}{\textbf{Method}} & \textbf{AI2D} & \textbf{ChartQA} & \textbf{DocVQA} & \textbf{MME} & \textbf{OCRBench} & \textbf{POPE} & \textbf{SQA} & \textbf{VQA\textsuperscript{Text}} & \multirow{2}{*}{\textbf{Avg}} \\
  & \textit{EM} & \textit{Relaxed} & \textit{Anls} & \textit{Score} & \textit{Acc} & \textit{Acc} & \textit{EM} & \textit{EM} &  \\
\midrule
Baseline & 82.48 & 83.68 & 94.90 & 1701 & 85.30 & 87.80 & 88.45 & 82.74 & 100.00 \\
\midrule
\multicolumn{10}{l}{\textit{Strategy 1: DPP + VisionSelector}} \\
DPP+VisionSelector & 79.70 & 73.36 & 93.02 & 1691 & 73.00 & 86.96 & 84.73 & 80.69 & 94.95 \\
\midrule
\multicolumn{10}{l}{\textit{Strategy 2: IDPruner (MMR Mechanism)}} \\
IDPruner ($\lambda=0.1$) & 78.08 & 67.36 & 87.94 & 1680 & 62.00 & 86.59 & 83.59 & 77.65 & 90.78 \\
IDPruner ($\lambda=0.3$) & 80.05 & 73.88 & 91.26 & 1671 & 70.50 & 86.91 & 84.09 & 79.69 & 94.10 \\
IDPruner ($\lambda=0.5$) & \textbf{80.51} & 74.32 & 93.16 & 1695 & \textbf{74.00} & 87.06 & \textbf{85.52} & 80.83 & \textbf{95.56} \\
IDPruner ($\lambda=0.7$) & 80.25 & 74.12 & 93.35 & \textbf{1710} & \textbf{74.00} & 87.07 & 85.13 & 80.75 & 95.56 \\
IDPruner ($\lambda=0.9$) & 79.66 & 72.72 & 93.29 & 1705 & 72.80 & 86.96 & 84.88 & 80.61 & 94.97 \\
\midrule
\multicolumn{10}{l}{\textit{Strategy 3: Naive Hybrid Selector}} \\
Hybrid ($\lambda=0.1$) & 78.79 & 66.52 & 86.90 & 1700 & 59.90 & 86.09 & 83.64 & 78.12 & 90.47 \\
Hybrid ($\lambda=0.3$) & 79.50 & 72.72 & 90.71 & 1704 & 62.60 & 86.56 & 83.34 & 79.91 & 92.73 \\
Hybrid ($\lambda=0.5$) & 79.95 & 74.36 & 92.20 & 1702 & 64.30 & \textbf{87.17} & 84.33 & 80.82 & 93.84 \\
Hybrid ($\lambda=0.7$) & 79.18 & \textbf{75.08} & 93.10 & 1680 & 66.40 & 86.63 & 84.83 & \textbf{80.94} & 94.10 \\
Hybrid ($\lambda=0.9$) & 79.31 & 73.84 & \textbf{93.42} & 1681 & 71.90 & 86.66 & 84.58 & 80.45 & 94.69 \\
\bottomrule
\end{tabular}
}
\caption{Ablation study of integration strategies on Qwen2.5-VL-7B-Instruct with 25\% token retention. We use VisionSelector as the base importance scorer. $\lambda$ controls the trade-off between importance and diversity.}
\label{tab:ablation_lambda}
\end{table*}

\noindent \textbf{Superiority of MMR Mechanism.}
The integration strategy plays a pivotal role in model performance.
As evidenced in Table \ref{tab:ablation_lambda}, \textbf{IDPruner} consistently outperforms the Naive Hybrid strategy across comparable $\lambda$ settings and also surpasses the DPP-based baseline.
The Naive Hybrid approach typically prioritizes tokens with the highest importance scores before applying Farthest Point Sampling (FPS) to enhance diversity.
However, this two-stage paradigm fails to address the inherent redundancy among high-importance tokens, resulting in a selected subset that lacks sufficient diversity.
In contrast, IDPruner employs a unified scoring mechanism that simultaneously manages importance and redundancy.
By dynamically penalizing semantically repetitive tokens during selection, our method achieves a more effective balance, thereby demonstrating superior robustness over heuristic hybrid strategies.

\noindent \textbf{Hyperparameter Selection.}
The hyperparameter $\lambda$ controls the balance between token importance and semantic diversity.
For IDPruner, the performance follows an inverted U-shape pattern, peaking at $\lambda=0.5$ with an average performance of \textbf{95.56\%}.
This confirms that setting $\lambda=0.5$ successfully strikes an optimal balance between token importance and semantic diversity, enabling IDPruner to leverage both properties for maximum performance.

\section{Limitations}

Despite the promising results achieved by IDPruner, we acknowledge certain limitations in this study.
First, constrained by computational resources, we have not yet evaluated our method on long-context video understanding benchmarks.
This restricts the comprehensive verification of our method's effectiveness in scenarios involving extremely long temporal sequences, thereby limiting the scope of applicable scenarios.
Second, due to time constraints, we did not conduct a fine-grained measurement or exhaustive search for the hyperparameter $\lambda$.
While the current settings demonstrate strong robustness, a more thorough optimization could potentially yield further performance improvements.

\section{Visualization}
\label{sec:visualization}

\begin{figure*}[t]
    \centering
    
    \begin{minipage}{0.24\textwidth}
        \centering
        \textbf{Original Image} \\ \vspace{2pt} 
        \includegraphics[width=\linewidth]{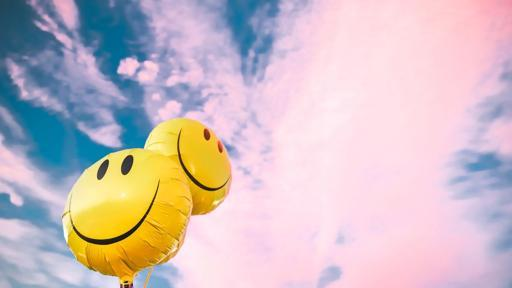} 
    \end{minipage}
    \hfill
    \begin{minipage}{0.24\textwidth}
        \centering
        \textbf{DivPrune} \\ \vspace{2pt}
        \includegraphics[width=\linewidth]{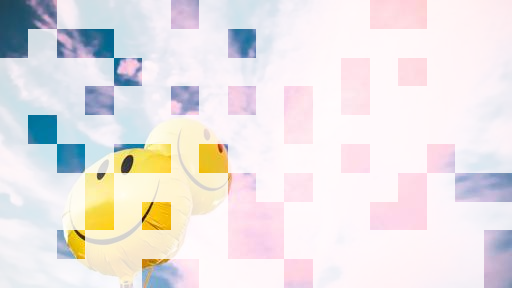} 
    \end{minipage}
    \hfill
    \begin{minipage}{0.24\textwidth}
        \centering
        \textbf{VisionSelector} \\ \vspace{2pt}
        \includegraphics[width=\linewidth]{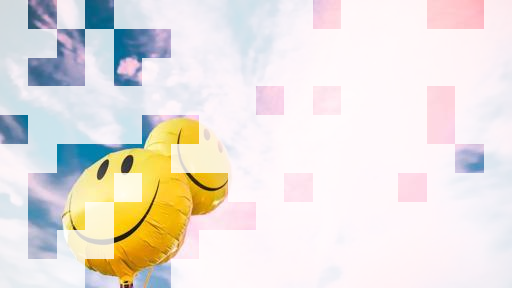} 
    \end{minipage}
    \hfill
    \begin{minipage}{0.24\textwidth}
        \centering
        \textbf{IDPruner (Ours)} \\ \vspace{2pt}
        \includegraphics[width=\linewidth]{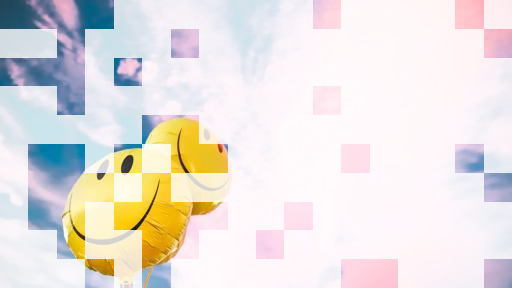} 
    \end{minipage}
    
    \vspace{4pt} 
    
    \begin{minipage}{0.24\textwidth}
        \centering
        \includegraphics[width=\linewidth]{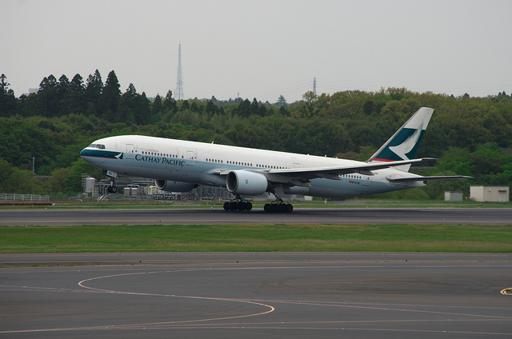} 
    \end{minipage}
    \hfill
    \begin{minipage}{0.24\textwidth}
        \centering
        \includegraphics[width=\linewidth]{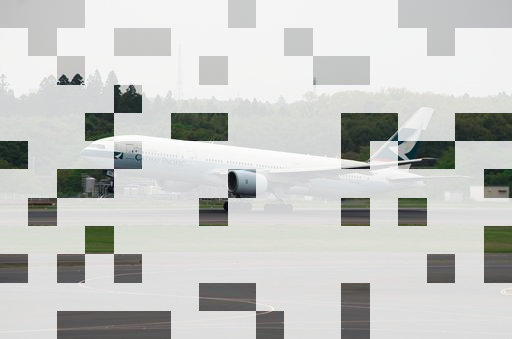} 
    \end{minipage}
    \hfill
    \begin{minipage}{0.24\textwidth}
        \centering
        \includegraphics[width=\linewidth]{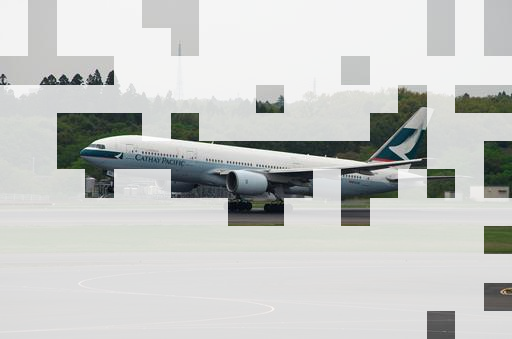} 
    \end{minipage}
    \hfill
    \begin{minipage}{0.24\textwidth}
        \centering
        \includegraphics[width=\linewidth]{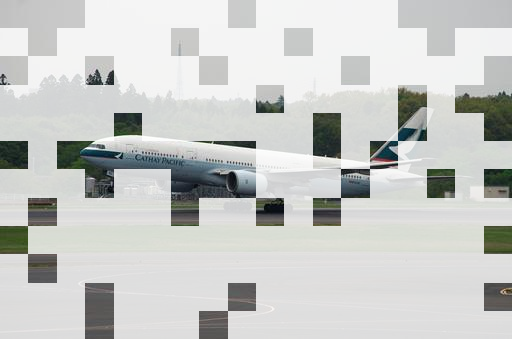} 
    \end{minipage}
    
    \vspace{4pt} 

    \begin{minipage}{0.24\textwidth}
        \centering
        \includegraphics[width=\linewidth]{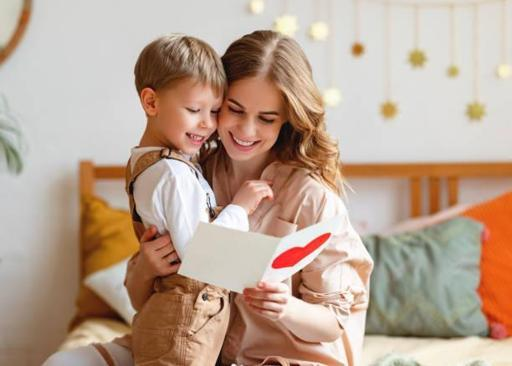} 
    \end{minipage}
    \hfill
    \begin{minipage}{0.24\textwidth}
        \centering
        \includegraphics[width=\linewidth]{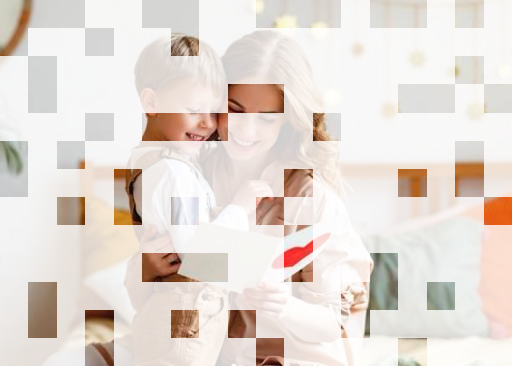} 
    \end{minipage}
    \hfill
    \begin{minipage}{0.24\textwidth}
        \centering
        \includegraphics[width=\linewidth]{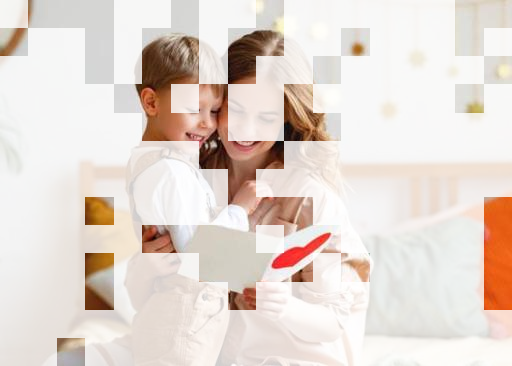} 
    \end{minipage}
    \hfill
    \begin{minipage}{0.24\textwidth}
        \centering
        \includegraphics[width=\linewidth]{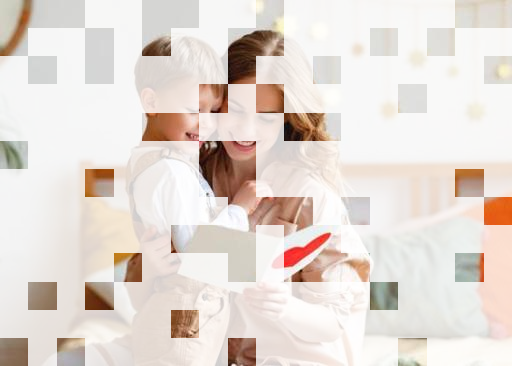} 
    \end{minipage}
    
    \vspace{4pt} 

    \begin{minipage}{0.24\textwidth}
        \centering
        \includegraphics[width=\linewidth]{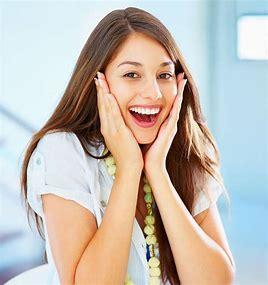} 
    \end{minipage}
    \hfill
    \begin{minipage}{0.24\textwidth}
        \centering
        \includegraphics[width=\linewidth]{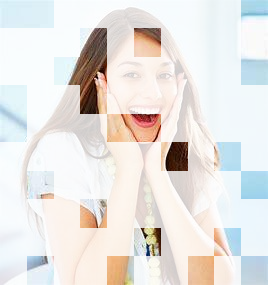} 
    \end{minipage}
    \hfill
    \begin{minipage}{0.24\textwidth}
        \centering
        \includegraphics[width=\linewidth]{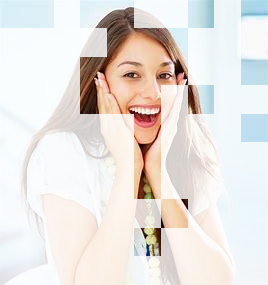} 
    \end{minipage}
    \hfill
    \begin{minipage}{0.24\textwidth}
        \centering
        \includegraphics[width=\linewidth]{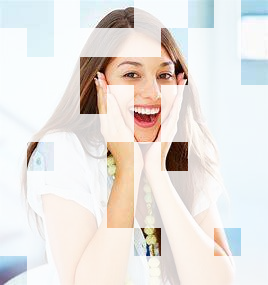} 
    \end{minipage}
    
    \caption{
        \textbf{Visualization of retained visual tokens across different samples from MMBench.}
        Columns from left to right: Original Image, DivPrune, VisionSelector, and IDPruner.
        \textbf{DivPrune} maintains global coverage but often neglects the semantic subject.
        \textbf{VisionSelector} clusters heavily on salient objects, resulting in redundancy and background loss.
        \textbf{IDPruner} achieves a superior balance, preserving intricate details of the subject while maintaining essential background context for global reasoning.
    }
    \label{fig:visualization_masks_all}
\end{figure*}

To intuitively understand how IDPruner harmonizes importance and diversity compared to existing approaches, we visualize the spatial distribution of retained visual tokens across multiple samples.
Figure \ref{fig:visualization_masks_all} presents a comparison of token selection masks under a 25\% retention ratio.
As consistently observed across diverse scenes, \textbf{DivPrune} tends to produce a uniform distribution, often overlooking semantic details. \textbf{VisionSelector} overly concentrates on foreground objects at the expense of background information coverage. In contrast, \textbf{IDPruner} successfully balances both, capturing salient features while maintaining essential background context necessary for global reasoning.

\section{Empirical Verification of Non-Negative Similarity}
\label{sec:similarity_analysis}

A potential concern regarding the MMR mechanism is the behavior of the redundancy penalty term, $(1 - \lambda) \cdot \text{Sim}(v_i, v_j)$.
If the cosine similarity $\text{Sim}(v_i, v_j)$ were to yield negative values (implying an angle $\theta > 90^\circ$ between feature vectors), the intended penalty would transform into a reward.

To address this validity concern, we empirically analyzed the geometric properties of the visual token space.
We randomly selected 100 images from the MMBench dataset and computed the pairwise angles between all visual tokens extracted from the \textbf{Qwen2.5-VL-7B-Instruct} model.

\begin{figure}[h]
    \centering
    \includegraphics[width=\linewidth]{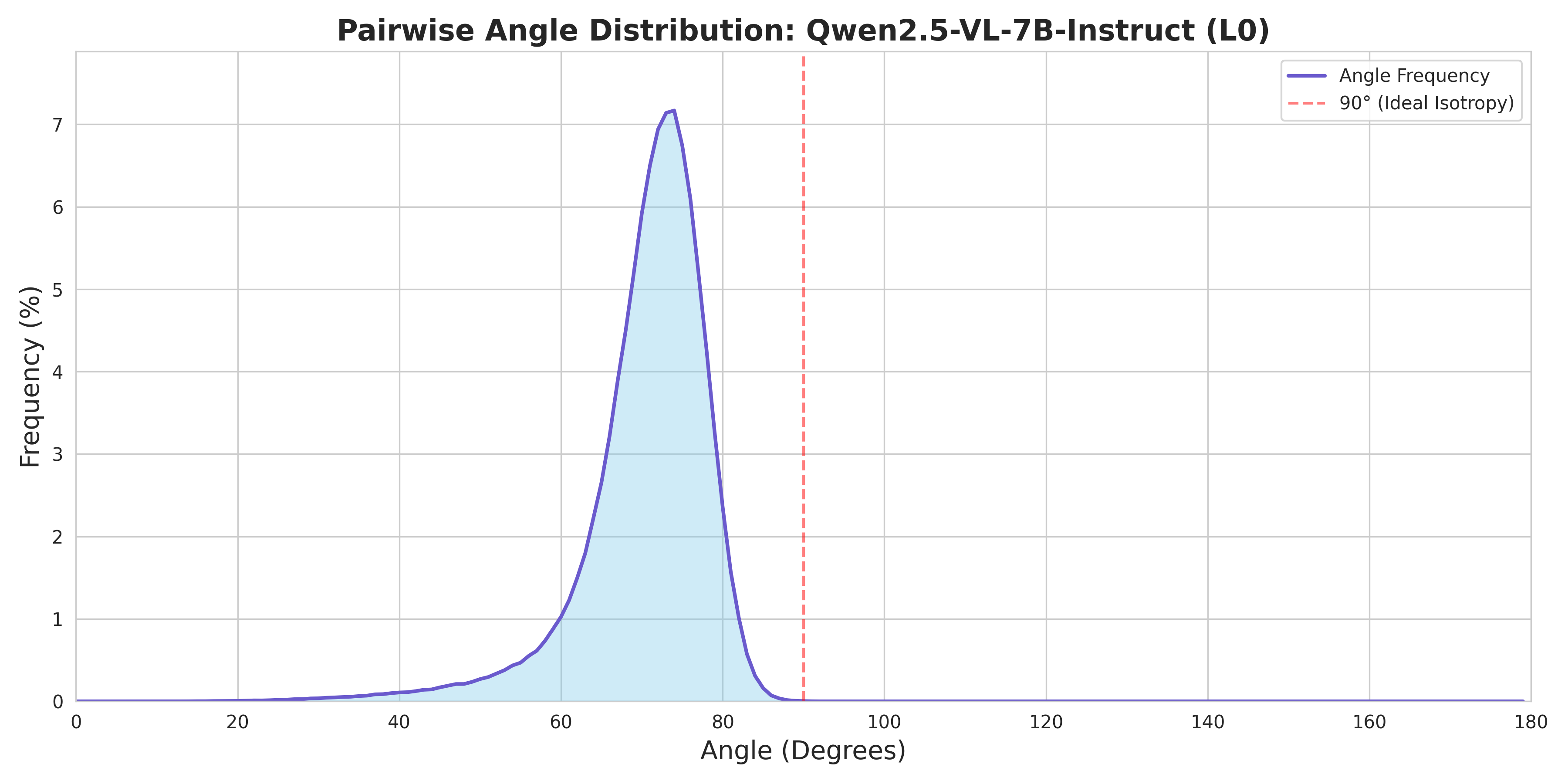} 
    \caption{
        \textbf{Distribution of pairwise angles between visual tokens.}
        We calculated the angles for all token pairs across 100 images from MMBench using Qwen2.5-VL-7B.
        The distribution is entirely concentrated within the acute angle range ($< 90^\circ$), peaking around $74^\circ$.
        The absence of obtuse angles ($> 90^\circ$, right of the red dashed line) guarantees that the cosine similarity metric remains strictly non-negative.
    }
    \label{fig:angle_distribution}
\end{figure}

As illustrated in Figure \ref{fig:angle_distribution}, the distribution of pairwise angles exhibits a distinct pattern.
The distribution is overwhelmingly concentrated within the range of $[0^\circ, 85^\circ]$, with a peak density at approximately $74^\circ$.
Crucially, there is zero probability mass beyond the $90^\circ$ threshold (indicated by the red dashed line).

Since $\text{Sim}(v_i, v_j) = \cos(\theta_{ij})$ and $\cos(\theta) \ge 0$ for all $\theta \in [0^\circ, 90^\circ]$, this empirical evidence confirms that all similarity scores in our framework are strictly non-negative.
Consequently, the term $(1 - \lambda) \cdot \text{Sim}(v_i, v_j)$ consistently functions as a redundancy penalty, validating the theoretical soundness of our IDPruner formulation.

\section{Statement on the Use of AI Assistants}
\label{sec:ai_statement}

In accordance with the ACL submission policies, we hereby declare the use of AI assistants in the preparation of this manuscript.
We utilized AI assistants for writing refinement, including grammar correction, vocabulary enhancement, and proofreading to improve readability.
We emphasize that all scientific claims, experimental designs, core concepts, and logical arguments presented in this work are the original contributions of the authors.
All AI-generated content was meticulously reviewed and verified by the authors to ensure accuracy and adherence to academic standards; the authors assume full responsibility for the content of this paper.

\label{sec:appendix}

\end{document}